\documentclass{ametsocV6.1}

\usepackage{array}
\usepackage{subcaption}
\title{Hydra-LSTM: A semi-shared Machine Learning architecture for prediction across Watersheds}

\authors{Karan Ruparell,\aff{a, b}\correspondingauthor{Karan Ruparell, k.ruparell2@pgr.reading.ac.uk} 
Robert J. Marks,\aff{a} 
Andy Wood,\aff{c} 
Kieran M. R. Hunt,\aff{a,d} 
Hannah L. Cloke,\aff{a, e} 
Christel Prudhomme,\aff{b} 
Florian Pappenberger,\aff{b} 
Matthew Chantry\aff{b}
}

\affiliation{\aff{a}{University of Reading, Department of Meteorology, Reading, UK}\\
\aff{b}{European Centre for Medium-Range Weather Forecasts (ECMWF), Reading, UK}\\
\aff{c}{National Center for Atmospheric Research (NCAR), Boulder, Colorado}\\
\aff{d}{National Centre for Atmospheric Science, Reading, UK}\\
\aff{e}{University of Reading, Department of Geography and Environmental Science, Reading, UK}
}

\abstract{Long Short Term Memory networks (LSTMs) are used to build single models that predict river discharge across many catchments. These models offer greater accuracy than models trained on each catchment independently if using the same data. However, the same data is rarely available for all catchments. This prevents the use of variables available only in some catchments, such as historic river discharge or upstream discharge. The only existing method that allows for optional variables requires all variables to be considered in the initial training of the model, limiting its transferability to new catchments. To address this limitation, we develop the Hydra-LSTM. The Hydra-LSTM processes variables used across all catchments and variables used in only some catchments separately to allow general training and use of catchment-specific data in individual catchments. The bulk of the model can be shared across catchments, maintaining the benefits of multi-catchment models to generalise, while also benefitting from the advantages of using bespoke data. We apply this methodology to 1 day-ahead river discharge prediction in the Western US, as next-day river discharge prediction is the first step towards prediction across longer time scales. We obtain state-of-the-art performance, generating more accurate median and quantile predictions than Multi-Catchment and Single-Catchment LSTMs while allowing local forecasters to easily introduce and remove variables from their prediction set. We test the ability of the Hydra-LSTM to incorporate catchment-specific data by introducing historical river discharge as a catchment-specific input, outperforming state-of-the-art models without needing to train an entirely new model.}

\begin{document}

%% Necessary!
\maketitle

%%%%%%%%%%%%%%%%%%%%%%%%%%%%%%%%%%%%%%%%%%%%%%%%%%%%%%%%%%%%%%%%%%%%%
% SIGNIFICANCE STATEMENT/CAPSULE SUMMARY
%%%%%%%%%%%%%%%%%%%%%%%%%%%%%%%%%%%%%%%%%%%%%%%%%%%%%%%%%%%%%%%%%%%%%
%
% If you are including an optional significance statement for a journal article or a required capsule summary for BAMS 
% (see www.ametsoc.org/ams/index.cfm/publications/authors/journal-and-bams-authors/formatting-and-manuscript-components for details), 
% please apply the necessary command as shown below:
%
% Significance Statement (all journals except BAMS)
%
%\statement
%	 Enter significance statement here, no more than 120 words. See \url{www.ametsoc.org/index.cfm/ams/publications/author-information/significance-statements/} for details.
%
%% Capsule (BAMS only)
%%
%\capsule
%       Enter BAMS capsule here, no more than 30 words. See \url{www.ametsoc.org/index.cfm/ams/publications/author-information/formatting-and-manuscript-components/#capsule} for details.
%
%% * * If using twocol mode, you will need to use the commands "twocolsig" and "twocolcapsule" in place of "sig" and "capsule"
%%      to ensure that the text box correctly spans across both columns.
%

%%%%%%%%%%%%%%%%%%%%%%%%%%%%%%%%%%%%%%%%%%%%%%%%%%%%%%%%%%%%%%%%%%%%%
% MAIN BODY OF PAPER
%%%%%%%%%%%%%%%%%%%%%%%%%%%%%%%%%%%%%%%%%%%%%%%%%%%%%%%%%%%%%%%%%%%%%
%

%% In all cases, if there is only one entry of this type within
%% the higher level heading, use the star form: 
%%
% \section{Section title}
% \subsection*{subsection}
% text...
% \section{Section title}

%vs

% \section{Section title}
% \subsection{subsection one}
% text...
% \subsection{subsection two}
% \section{Section title}

\section{Background} \label{Background}

Accurate river discharge forecasts are vital for catchment managers to decide how much water to extract for agricultural usage, the production of hydroelectricity, or any actions to take to maintain the river's biodiversity. The needs of catchment managers and the relevant variables for forecasts can vary greatly between catchments (Li and Razavi 2024), making it vital that models are readily adaptable and expandable in a low-cost way to take advantage of local domain expertise (\cite{fleming2019machine, fleming2021assessing}). In the current state of the art of machine learning for river discharge forecasting, if a variable is essential in a catchment but is not in the list of variables the model was initially trained for, it cannot be used without retraining the entire model. This creates a conflict in the training of machine learning models, as individual models perform better when trained across a range of catchments  (Kratzert et al. 2019), but by doing so, only variables available in all catchments can be used, potentially excluding data that may improve forecasts.

\cite{kratzert2019toward, kratzert2023never} have popularised the use of Long Short Term Memory networks in river discharge prediction, a form of recurrent neural network. They showed both that LSTMs have the potential to outperform physical models (2018), and that training an LSTM across multiple catchments can outperform a collection of LSTMs each trained on an individual catchment. Having a model that is trained across multiple catchments has a wide array of benefits, allowing forecasters to use models that would be too expensive to train themselves and making the model more robust to changes in climatic conditions as the model has familiarity with a deeper range of climatic conditions by being trained across catchments (Kratzert et al. 2023). However, LSTMs require the same data to be available and used in every application of a given pre-trained model, which is impractical in a multi-catchment setting without severely limiting the usability of these models. For example, recent river discharge is important for predicting future river discharge. However, in many catchments river discharge observations do not exist or are only available with a large time delay. Under the method, to train Multi-Catchment LSTMs that are usable at river gauges that do have river discharge available as a potential input and river gauges that do not, you have to treat river discharge as unusable across all catchments. 

This forces forecasters into a choice: should they choose a model that lets them benefit from knowledge gained from other catchments, or should they choose a model tailored to their catchment? The latter allows them to include information like historical river discharge, local forecast predictions, additional soil measurements, upstream observations, and other variables that are important in many catchments but only sometimes available. This underpins the need for a model like the Hydra-LSTM, to provide a third choice: a model that can take advantage of increased exposure to catchment variability while preserving the importance of bespoke catchment knowledge.

The only current attempt to address this need (Nearing et al. 2023) allows variables to be input into a single general model alongside a corresponding flag variable that is 1 if the variable is available and 0 if not. Missing variables are replaced by some default value, with a 0 in the corresponding flag variable. This approach allows the same model to be used on different catchments, even if they have other data available. However, it is unusable for variables not already defined in the parameter set or for bespoke variables, such as information from upstream gauges.

\begin{figure}[h]
    \centering
    \includegraphics[width=0.9\textwidth]{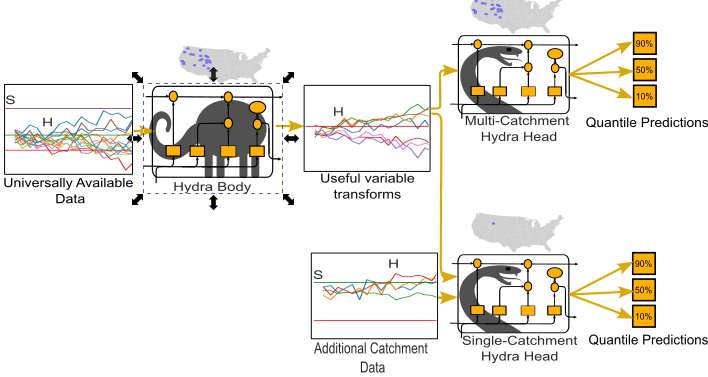}
    \caption{Diagram of Hydra Model Architecture. The leftmost box plots the time series data available at all catchments, including historical data, forecast data, and static catchment attributes. An encoding LSTM processes these, dubbed the Hydra Body, which produces a lower dimensional encoding of the information. If no further data is available, this encoding is passed to the Multi-Catchment Head, an LSTM that transforms the encoding to quantile discharge predictions. However, if further information is available for that catchment, it is passed to a Single-Catchment Head alongside the additional time series data, which are then combined to produce quantile discharge predictions. }
    \label{fig:Hydra_Diagram}
\end{figure}

Here, we design a model that can combine the ability to train on many catchments with the flexibility to use whatever relevant hydrological information is available. We call this model the Hydra-LSTM (Figure \ref{fig:Hydra_Diagram}). The Hydra-LSTM uses an initial encoding LSTM, called the Hydra Body, to use variables available across all catchments, for example, globally available reanalysis data sets such as ERA5 (Hersbach et al. 2020) The Hydra Body transforms them into a set of encoding variables that are more directly useful to river discharge prediction. The Hydra Body outputs are then combined with one of the Hydra Heads, each an LSTM itself, being passed to either a Multi-Catchment Head trained across all catchments or a Single-Catchment Head trained only for the corresponding catchment that is capable of accepting catchment-specific variables. The predictions produced by the Hydra Heads are the 10\% and 90\% quantiles of river discharge for the next day. This gives a measure of the uncertainty of the model predictions, which catchment managers can then consider before taking actions based on the forecasts. Our model can be trained to predict these using the quantile loss in Equation \ref{eq:quantile_loss}. 

The Single-Catchment Head might then take the outputs of the Body alongside a time series of recent river discharge observations from that river gauge and an upstream river gauge that the forecaster has decided is particularly influential to the river flow at this river gauge. The Single-Catchment Head would then predict the next day's river discharge at that river gauge, which has been tuned specifically to that river gauge.

\section{Experimental Design}

We test the Hydra-LSTM for 1-day-ahead river discharge prediction, as this is the foundation for prediction over longer lead times, and a number of other LSTM-based models have shown the potential of machine learning models at this lead time (\cite{kratzert2018rainfall, hunt2022using, nearing2023ai}). We focus on creating forecasts of the 10\%, 50\%, and 90\% quantile thresholds of next-day river discharge, as it is useful for water forecasters to understand the range of uncertainty in predictions (Wang et al. 2019), , and quantile estimation has been successfully applied in hydrological forecasting previously (\cite{jahangir2023quantile, koenker2005quantile}). 

For this task, there are two requirements for the Hydra-LSTM to be beneficial. Firstly, when the same data are available, it should perform just as well as other state-of-the-art models. The main benefit of multi-catchment models, such as the Hydra-LSTM, is their ability to be used out of the box, so a forecaster should be able to use the Hydra-LSTM in this setting without having to pay a penalty compared to the skill they could have achieved if they used another model. Secondly, the Hydra-LSTM should benefit from including catchment-specific data in an additional Single-Catchment Head. The unique advantage of the Hydra-LSTM is in its ability to include additional data not considered in the design of the Hydra Body and multi-catchment Head, and so in order for the Hydra-LSTM to be beneficial, the Single-Catchment Head should be a useful means of adding additional data in a way that improves the model's performance. 

To test these criteria, we perform two experiments. In the first experiment, we evaluate the Hydra-LSTM Body and Multi-Catchment Head in a setting where historical river discharge is not operationally available and compare it to other state-of-the-art models, described in section \ref{Models}. In this experiment none of the models are given access to historical river discharge as a prediction input, however it is used for training the model. In the second experiment, we provide all models with river discharge as an input. We provide river discharge into the Hydra-LSTM as an additional data source in Single-Catchment Heads for each catchment. If historic river discharge added this way can be used just as well as if it were fed as an input in the multi-catchment setting, then we expect the performance of the Hydra-LSTM to be at least as good as the performance of the benchmarks. This would mean that we can add additional data to our Hydra-LSTM without retraining the entire model. This is not true for existing LSTMs used in river discharge prediction, where adding additional data to any of the machine learning benchmarks would require retraining us to retrain entire multi-catchment model with that variable present.

\subsection{Performance Benchmarks} \label{Benchmarks}
We compare our Hydra-LSTM with four different machine learning approaches already used for predicting daily river discharge, all LSTMs (\cite{kratzert2018rainfall, kratzert2019toward, nearing2023ai}). These models are the current best approaches to using a machine learning model for river discharge prediction in catchments, and so are alternative models that might otherwise be used by a forecaster that could use the Hydra-LSTM.

The first of these modelling approaches is to use Single-Catchment LSTMs, training separate LSTMs for each catchment. The architecture is replicated as described in \cite{kratzert2018rainfall}. This allows all available data in each catchment to be used in training but means that each model is not trained on more than one catchment. This means it is less exposed to extremes, which can decrease its ability to extrapolate. This requires individual forecasters to set up the model from scratch, adding additional complexity in usability and means that the model has less training data to use, potentially decreasing performance and generalisability to different hydrological conditions. This is an approach that is likely to be used when a forecaster has some specific feature they would wish to use in their model that is not present in any out-of-the-box multi-catchment model and has enough data in their own catchment to train a suitable model. We are interested in seeing whether or not the Single-Catchment Head of the Hydra-LSTM can offer results comparable to or even outperform these Single-Catchment LSTMs. 

Secondly, we train two Multi-Catchment LSTMs, one with river discharge as a predictor and one without. The implementation follows the method described in \cite{kratzert2019toward}. These LSTMs are trained across all catchments in our dataset. The decision whether or not to use the Multi-Catchment LSTM, and whether to use it with or without river discharge usually depends on whether or not you have river discharge available operationally as an input, as many catchments will not. Multi-Catchment LSTM setups do not allow for any differences in data availability between uses, either always requiring a variable or having no way of using it. It has been shown that with enough data across catchments to train on, these models can outperform Single-Catchment LSTMs and are especially useful in cases when the gauge record is too short to train a Single-Catchment LSTM. They are also useful when no additional data would allow the forecaster to benefit from a tailored approach. Here, we are interested in whether or not the Hydra-LSTM can perform similarly to the Multi-Catchment LSTM trained without river discharge when it also does not use river discharge, i.e. through the Multi-Catchment Head. We also compare the performance of the Single-Catchment Head, which has river discharge as an additional input, with that of the Multi-Catchment LSTM, which has river discharge as an input. If it performs similarly in the setting where river discharge is available, then we will have shown that there is no relevant loss in skill by providing additional information through an additional head instead of retraining an entirely separate Multi-Catchment LSTM with river discharge as an input. This would offer more flexibility in variable choice, allowing forecasters to train Single-Catchment Heads onto the Hydra-LSTM using just the data at their catchment instead of having to train a multi-catchment model. 

The final model we compare against can tackle data disparity between catchments, which we call a Flag LSTM and is a method used in \cite{nearing2023ai}. This model has different variables as potential inputs and, for each of these variables, a corresponding flag time series denoting whether the variable is available for a given day. If the data is missing, the flag time series contains a zero; otherwise, it contains a one. The corresponding variable time series has some placeholder numbers for days when they are not available. In our tests, river discharge is the additional variable that may be unavailable. This model is able to make predictions both when river discharge is and is not available, but it does not allow for any additional variables to be introduced after the initial design of the model without retraining the entire model. Our Hydra-LSTM, on the other hand, has a Single-Catchment Head that allows additional variables to be introduced by training only a smaller new section of the model, and so it is far easier to introduce new variables.  We train the Flag LSTM, using a flag to specify whether or not river discharge is available at a catchment. In training, 50\% of training examples are without river discharge. When comparing with this model, we wish to see how the method of adding additional data through an additional model head compares to using a binary flag to introduce potentially missing data. Even though the Hydra-LSTM is more flexible, allowing additional data to be used even when it is not considered at the model development stage, the additional flexibility the Flag LSTM provides when compared to the other benchmarks may be satisfactory in many cases if it performs significantly better than the Hydra-LSTM.

\begin{figure}[h]
    \centering
    \includegraphics[width=\textwidth]{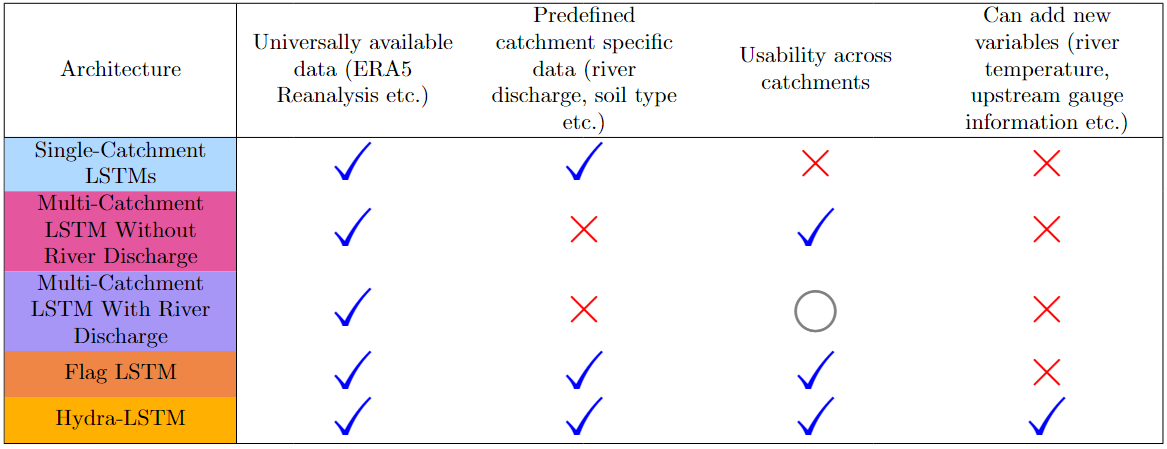} % Adjust path and filename accordingly
    \caption{Comparison of different LSTMs for hydrological modelling regarding their data requirements, usability across catchments, and flexibility in adding new variables. The architectures include Single-Catchment LSTMs, Multi-Catchment LSTM without River Discharge, Multi-Catchment LSTM with River Discharge, Flag LSTM, and Hydra-LSTM. The checkmarks indicate the presence of a feature, the crosses indicate the absence of a feature, and the circles indicate partial usability.}
    \label{fig:Models}
\end{figure}

\section{Hydra-LSTM Architecture} \label{Models}

\subsection*{Hydra-LSTM Architecture}

The Hydra Model consists primarily of three blocks: the Hydra Body, the Multi-Catchment Head, and a Single-Catchment Head for each catchment wishing to incorporate additional data. In our experiments, each of these model blocks are LSTMs. LSTMs are a form of recurrent neural network, and often used when making predictions with temporal inputs. However, this general architecture could be adapted to use an array of different architectures. As our implementation of the Hydra Model comprises LSTMs, it takes in data as a vector of time series. Static catchment variables, such as soil type or rock type, are then passed to a model as unchanging time series.  

The Hydra-Body takes as inputs time-series data that is available across all catchments, such as ERA5 reanalysis precipitation or catchment area, and transforms these into smaller and more informative time-series of encoded variables. If a water forecaster does not wish to use a bespoke Single-Catchment Head for their catchment, this time series of encodings are passed to the Multi-Catchment Head as inputs. The Multi-Catchment Head then returns a set of three predictions, predicting the 10\%, 50\%, and 90\% quantile thresholds of the next day's river discharge. 

If a water forecaster does wish to use a Single-Catchment Head, however, then the outputted time series from the Hydra Body can be concatenated with additional catchment data and then be passed onto their catchment Single-Catchment Head. The Single-Catchment Head then outputs predictions of the 10\%, 50\%, and 90\% quantile thresholds of the next day's river discharge, just as the Multi-Catchment Head would. In our experiments, we assume that river discharge is not always operationally available, and so do not include it as inputs into the Hydra Body. We also train a different Single-Catchment Head for each catchment, with river discharge as an additional input that is concatenated to the outputs of the Hydra Body. We also further analyse prediction quality at the Green River Below
Howard A Hanson Dam gauge.

\newpage 

\section{Data}

The Water Supply Forecast Rodeo was a competition held by the US Bureau of Reclamation in 2023 focused on predicting river discharge across 27 catchments in the western US for primarily agricultural and hydro-electrical purposes (DrivenData 2024). Because this work focuses on forecasting daily river discharge, we focus on 18 of those catchments for which daily river discharge observations are available. These catchments cover a wide range of different climatological conditions (Table \ref{tab:variable_range}) and static physical characteristics (Table \ref{tab:catchment_means}).

Historical observations from the US Geological Survey and reanalysis data from the ECMWF Reanalysis v5 Land (ERA5-Land) are used as input to the models with 25 variables used in total \cite{hersbach2020era5} (Table \ref{tab:variable_range}).  Daily river discharge observations are extracted from the \cite{usgs2024}.  ERA5-Land data is taken for the study region from October to July from 2001 - 2023  in hourly and 6 hourly intervals at a spatial resolution of 9km. Historical daily river discharge observations were only used in some catchments to test how introducing river discharge in a Single-Catchment Head in the Hydra-LSTM compares to introducing it in the benchmark models. Catchment attributes are taken as static for the purpose of this study and are taken from the BasinAtlas dataset (Linke et al. 2019)

\begin{figure}[h]
    \centering
    \includegraphics[width = 0.7\textwidth]{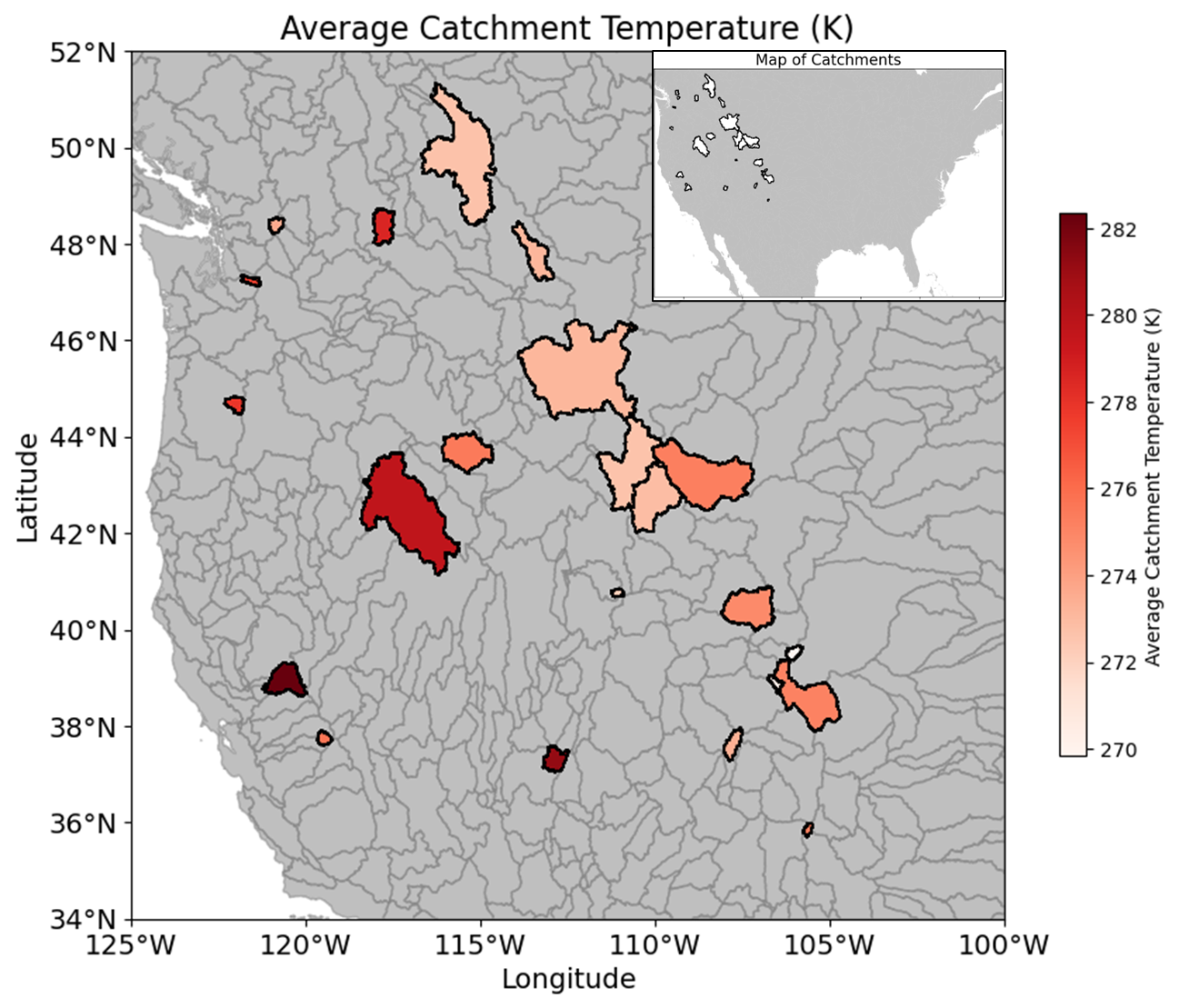}
    \caption{Plot of catchment sites evaluated in this study, and the mean annual temperature over each catchment in Kelvin. All catchments are located in the Western US, and the plot shows the western US with an inlet showing a map of US as a whole}
    \label{fig:Western_US_Diagram}
\end{figure}

% \newpage
\clearpage

\begin{table}[h]
\centering
\caption{Summary statistics of each variable used in training and their range across catchments.}

\begin{subtable}{1\textwidth}
    \centering
    \resizebox{\textwidth}{!}{%
    \begin{tabular}{|>{\centering\arraybackslash}p{0.25\linewidth}|>{\centering\arraybackslash}p{0.19\linewidth}|>{\centering\arraybackslash}p{0.19\linewidth}|>{\centering\arraybackslash}p{0.2\linewidth}|>{\centering\arraybackslash}p{0.14\linewidth}|}
        \hline
        Variable & Minimum of Catchment Means & Median of Catchment Means & Maximum of Catchment Means  &Source\\ \hline
        Precipitation (mm/day) & 0.011 & 0.021 & 0.070  &ERA5-Land\\ \hline 
        % Runoff & 0.000 & 0.009 & 0.055 \\ \hline 
        Evaporation (mm/day) & 0.87 & 1.2 & 1.5  &ERA5-Land \\ \hline 
        2m Temperature (K) & 272.3 & 277.4 & 284.5  &ERA5-Land \\ \hline 
        Snow Depth Water Equivalent (m) & 0.0127 & 0.0891 & 0.573  &ERA5-Land \\ \hline 
        Soil Water Volume (m³) & 0.14 & 0.30 & 0.35  &ERA5-Land \\ \hline
        River Discharge (m³/s) & 0.0955 & 1.3696 & 21.372  &USGS \\ \hline 
        10m U component of wind (m/s) & -0.09 & 0.67 & 2.14  &ERA5-Land \\ \hline 
        10m V component of wind (m/s) & -0.25 & 0.39 & 1.11  &ERA5-Land \\ \hline 
        Surface Net Solar Radiation (J/m$^{2}$) & 1.1e7 & 1.4e7 & 1.9e7  &ERA5-Land \\ \hline 
        Surface Net Thermal Radiation (J/m$^{2}$) & -8.8e6  & -6.7e6 & -4.3e6  &ERA5-Land \\ \hline 

    \end{tabular}
    }
    \caption{Time series variables}
\end{subtable}

\vspace{1em} 

\begin{subtable}{1\textwidth}
    \centering
    \resizebox{\textwidth}{!}{%
    \begin{tabular}{|>{\centering\arraybackslash}p{0.25\linewidth}|>{\centering\arraybackslash}p{0.19\linewidth}|>{\centering\arraybackslash}p{0.19\linewidth}|>{\centering\arraybackslash}p{0.2\linewidth}|>{\centering\arraybackslash}p{0.14\linewidth}|}
        \hline
        Variable & Minimum of Catchment Means & Median of Catchment Means & Maximum of Catchment Means  &Source\\ \hline
        Gauge Elevation (m) & 1650 & 3670 & 4320  &US Bureau of Reclamation \\ \hline 
        Area (km$^{2}$) & 420 & 3580 & 38010  &US Bureau of Reclamation  \\ \hline 
        Average slope ($\circ$) & 36 & 144 & 276  &BasinAtlas \\ \hline
        Mean annual temperature (K) & 271.2 & 275.9 & 282.0  &BasinAtlas \\ \hline 
        Climate Moisture Index & -70 & -18 & 65  &BasinAtlas \\ \hline 
        Inundation Extent (\%) & 0.0 & 1.5 & 83.0  &BasinAtlas \\ \hline
        Wetland Extent (\%) & 0.0 & 2.5 & 19.5  &BasinAtlas \\ \hline 
        Permafrost Extent (\%) & 0.0 & 0.5 & 16.6  &BasinAtlas \\ \hline        
        Snow cover extent (\%) & 16.7 & 42.6 & 54.5  &BasinAtlas \\ \hline 
        Degree of Regulation & 0.0 & 48.3 & 817.5  &BasinAtlas \\ \hline 
        Lake Area (\%) & 0.0 & 5.9 & 31.6  &BasinAtlas \\ \hline
        Grassland (\%) & 0.0 & 0.4 & 24.6  &BasinAtlas \\ \hline
        Forest (\%) & 0.0 & 7.6 & 86.9  &BasinAtlas \\ \hline 
        Cropland (\%) & 8.7 & 82.7 & 100.0  &BasinAtlas \\ \hline 
        Shrubland (\%) & 0.00 & 2.1 & 15.6  &BasinAtlas \\ \hline 
    \end{tabular}
    }
    \caption{Static catchment attributes}
\end{subtable}

\label{tab:variable_range}
\end{table}

\clearpage

\begin{table}[h!]
\centering
\caption{Summary statistics of four key catchment attributes for each catchment in the study. All summaries are taken from the ERA5 deterministic forecast from 2000-2023, with Gauge Elevation provided by the US Bureau of Reclamation metadata}
\label{tab:catchment_means}
\resizebox{\textwidth}{!}{%
\begin{tabular}{|>{\centering\arraybackslash}m{0.11\linewidth}|>{\centering\arraybackslash}m{0.31\linewidth}|>{\centering\arraybackslash}m{0.11\linewidth}|>{\centering\arraybackslash}m{0.12\linewidth}|>{\centering\arraybackslash}m{0.13\linewidth}|>{\centering\arraybackslash}m{0.1\linewidth}|}
\hline
\textbf{USGS id} & \textbf{USGS name}                           & \textbf{Mean Precipitation (mm/day)} & \textbf{Mean Evaporation (mm/day)} & \textbf{Mean Temperature (K)} & \textbf{Gauge Elevation (m)} \\ \hline
12362500         & S F Flathead River nr Columbia Falls MT      &  2.49                                          &  1.19                &   276.53                        &  930
                          \\ \hline
13037500         & Snake River NR Heise ID                      &  2.17                                          &  1.21                &   275.61                     &   1,530 
                          \\ \hline
7099400          & Arkansas River above Pueblo CO               &  1.44                                          &  1.28                &   277.02                      &   1,450 
                           \\ \hline
6054500          & Missouri River at Toston MT                  &  1.66                                          & 1.28                 &    276.20                     &    1,190 
                          \\ \hline
9361500          & Animas River at Durango, CO                  &  2.26                                          &   1.33              &   275.63                        &   1,980 
                           \\ \hline
9251000          & Yampa River Near Maybell, CO                 &  1.68                                          &  1.26                &  277.84                         &   1,800 
                           \\ \hline
12301933         & Kootenai River BL Libby Dam NR Libby MT      & 2.18                                          &   1.08               &   275.26                        &  640 
                            \\ \hline
13202000         & Boise River NR Boise ID                      &  1.91                                          &   1.23               &     278.80                       &   860 
                           \\ \hline
12105900         & Green River Below Howard A Hanson Dam, WA    & 4.68                                           &   1.41               &   279.41                        &   300 
                           \\ \hline
9109000          & Taylor River Below Taylor Park Reservoir, CO &  1.67                                          &  1.11                &  272.90                         &   2,800 
                           \\ \hline
9050700          & Blue River Below Dillon, CO                  & 1.90                                            &   1.16               &   272.66                         &   2,670 
                           \\ \hline
9211150          & Fontenelle Reservoir Near Fontenelle, WY     & 1.54                                            &    1.08              &   275.90                        &    1,980 
                          \\ \hline
10128500         & Weber River Near Oakley, UT                  &  1.93                                         &  1.43                &   276.59                        &   2,020
                           \\ \hline
11251000         & San Joaquin R BL Friant CA                   & 3.17                                           &  1.39                &  284.51                         &   90 
                           \\ \hline
11266500         & Merced R A Pohono Bridge NR Yosemite CA      &  2.93                                          &  1.18               &    278.68                        &    1,180 
                          \\ \hline
12409000         & Colville R at Kettle Falls, WA               & 1.75                                        &   1.31               &    281.41                       &   430
                           \\ \hline
12451000         & Stehekin River at Stehekin, WA               &   4.54                                         &   0.98               &     275.74                      &     340
                         \\ \hline
14181500         & North Santiam River at Niagara OR            & 4.39                                            &   1.26               &  280.50                         &  330
                            \\ \hline
9406000          & Virgin River at Virgin, UT                   & 1.15                                            &   0.87               &  283.39                         &   1,070 
                           \\ \hline
6259000          & Wind River Below Boysen Reservoir, WY        & 1.29                                            &    1.03              &  277.81                         &   1,400 
                           \\ \hline
8378500          & Pecos River Near Pecos, NM                   & 2.10                                           &   1.50           &   279.21                        &   2,290 
                           \\ \hline
13183000         & Owyhee River Below Owyhee Dam                & 1.03                                           &   0.88               &   282.43                        &     710
                         \\ \hline
\end{tabular}
}
\end{table}

\clearpage

\section{Training}

In order to train a single model to create predictions, $\hat{y}$ of different quantile thresholds $\tau$ of next-day river discharge, we need a loss function that is minimised when the model predicts the correct thresholds. The quantile loss, Equation \ref{eq:quantile_loss}, is an ideal function for this \cite{koenker2005quantile}. A generalisation of the Mean Absolute Error (MAE), the $\tau_{th}$ quantile loss $L_{\tau}$ can be proven to be minimised only by predicting the $\tau_{th}$ quantile of the distribution associated with the predictand, conditional on the available predictors. This means that applying the $10\%$ quantile loss to the model's prediction for the 10\% quantile threshold $\hat{y}_{10}$ results in the loss being minimised if and only if it consistently predicts the value of river discharge $y_{t}$ that has a 10\% chance of being exceeded given the data available to the model.  The loss is shown in Equation \ref{eq:quantile_loss}, where $y$ is the observed river discharge for the day the model is trying to make a prediction for, and the $\hat{y}$ is the models' corresponding prediction.

\begin{equation}\label{eq:quantile_loss}
L_{\tau}(y, \hat{y}) = 
\begin{cases}
(\tau - 1)(y - \hat{y}), & \text{if } y < \hat{y}, \\
\tau(y - \hat{y}), & \text{if } y \geq \hat{y}.
\end{cases}
\end{equation}

We calculate the loss using our model's corresponding quantile loss function for each threshold prediction. We then sum these losses to get a total loss for the set of quantiles, $L_{tot}$, in Equation \ref{eq:cumulative_quantile_loss}.  This equally weights each of the thresholds in the loss, which is also the weighting chosen in the Water Supply Forecast Rodeo \cite{drivendata2024reclamation}. 

\begin{equation} \label{eq:cumulative_quantile_loss}
    L_{tot}(y, \hat{y}) = \frac{1}{3}\left(L_{0.1}(y, \hat{y}_{0.1}) + L_{0.5}(y, \hat{y}_{0.5}) + L_{0.9}(y, \hat{y}_{0.9})\right)
\end{equation}

We normalise $L_{tot}$ by dividing it by the corresponding loss that would be obtained by a model that predicts the climatological quantile thresholds at each catchment, $L_{clim}$, which is analogous to the normalising factor in the Nash Sutcliffe Efficiency (NSE). This is done so that the loss does not consider the size of the catchment in determining the magnitude of the error but rather how much the model improves upon the simple benchmark of climatology, in terms of the percentage reduction of error. The climatology model is defined so that it predicts the 10\% quantile threshold for a given year on July 1st to be the lowest value that only 10\% of July 1sts in the historic dataset have exceeded.  We can then use this to define a Cumulative Quantile Efficiency Score (CQES), as 1 - $\frac{L_{tot}}{L_{clim}}$. A CQES value of 0 would respond to a model that is as skillful as climatology, and a CQES value of 1 would respond to a model that perfectly forecasts the observed values

\begin{equation} \label{eq: Cumulative Quantile Efficiency Score}
    CQES(y, \hat{y}) = 1 - \frac{L_{tot}(y, \hat{y})}{L_{clim}(y, y_{clim})} 
\end{equation}

As we wish the Hydra-LSTM to be useful to operational hydrologists, providing context into the methods by which machine learning models are trained is useful. To train machine learning models, we perform many stochastic updates to the model's parameters to minimise the loss of the model's predictions over a subset, hereafter `batch' of the examples in the training set. The gradient of the loss with respect to the model parameters is used in a process called Stochastic Gradient Descent. We use a variation of this method that considers the gradients at previous batches, known as the ADAM optimiser \cite{kingma2014adam}. This is a form of parameter optimization.

\begin{figure}[t]
    \centering
    \includegraphics[width=0.8\textwidth]{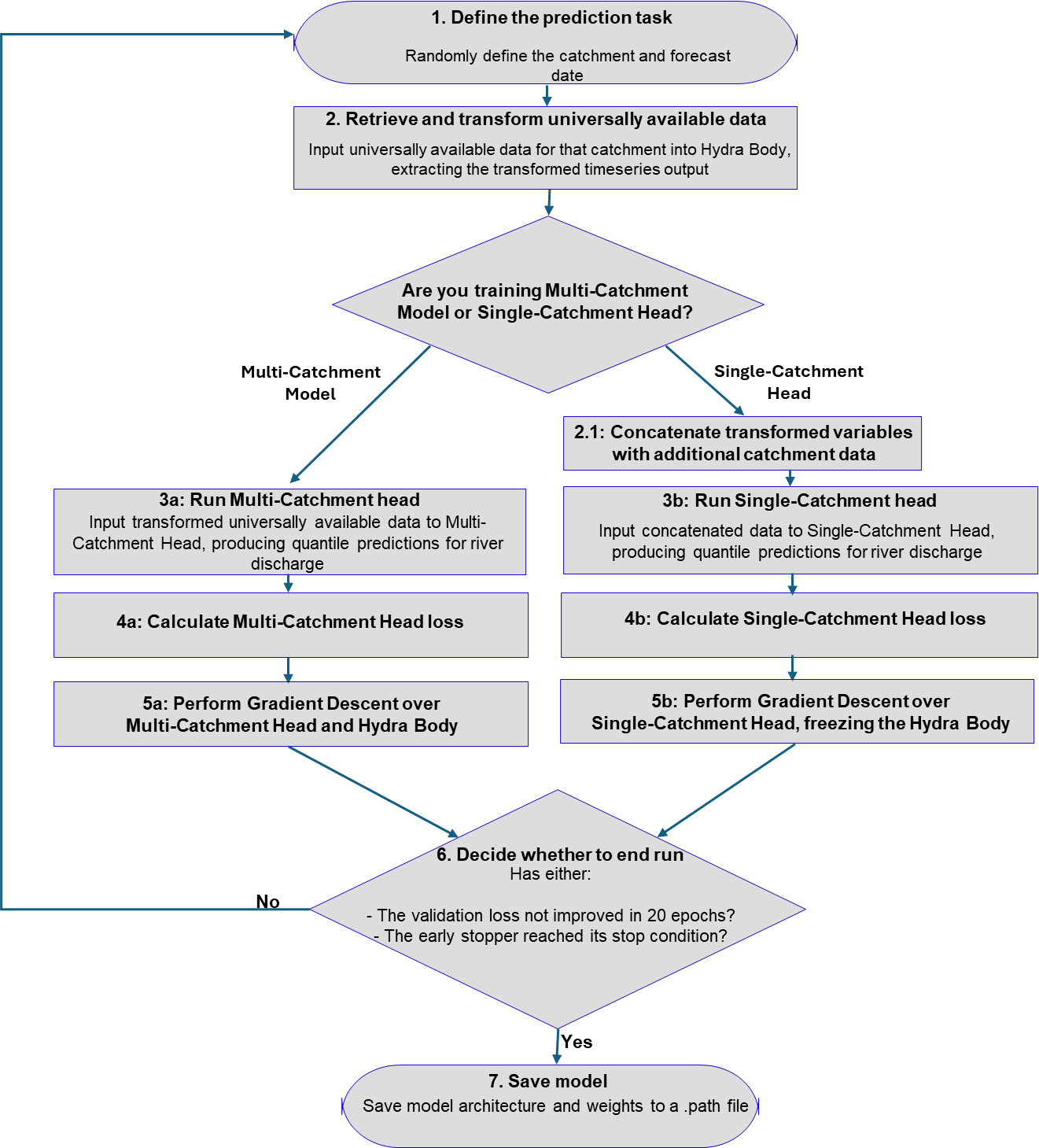}
    \caption{Training flow chart for Hydra-LSTM. The Multi-Catchment Head and Hydra Body are trained initially, and then the Single-Catchment Heads can be trained using the outputs from the Hydra Body as some of its inputs}
    \label{fig:Training_Diagram}
\end{figure}

Each training example is defined by the date a forecast is made and the catchment it is made for. In each batch, we randomly decided which catchment to make predictions for and randomly sampled potential forecast dates from which the prediction was made without replacement, step 1 in diagram \ref{fig:Training_Diagram}. For these catchments and dates, the relevant data is extracted in step 2, and fed into the model being trained on, outputting predictions for the different quantile thresholds in each training example in step 3. The predictions and actual observed values are then input into the loss function $L_{prop}$, step 4, and gradient descent is performed to update the model parameters, step 5. An epoch is complete when all forecast dates in the training set have been trained on, and we compute the loss for each epoch on a validation set. An early stopping check then takes the validation loss and decides whether or not to end the training procedure stopping the training loop if a new minimum validation loss has not been reached in the last 20 epochs, step 6. The early stopper does this depending on whether the maximum number of epochs has been reached or if the validation loss has started to plateau.  If the early stopper does end the training run, the model and its parameter weights are saved to be used later. 

For each model, we train it on a subset of 19 years and choose 2 years to validate. All multi-catchment models are trained on all catchments. The performance reported in the results section is then computed on a final year withheld from the training and validation set. We do this with 11 different potential test years, a process known as 11-fold cross-validation, or leave-one-out validation applied 11 times. 

For the Hydra-LSTM in particular, the Hydra Body and Multi-Catchment Head are trained in tandem, and then the Single-Catchment Heads are trained separately using the trained Hydra Body. This is because in operational usage we expect a Hydra Body and Head to be made initially, and then individual forecasters would train their own Single-Catchment Heads to incorporate their own additional data. This means that the Single-Catchment Head will have been trained separately from the Multi-Catchment parts, and so we want to test whether or not it will be able to satisfactorily combine the information from the Hydra Body with its additional data to make its prediction. The training of the Hydra-LSTM is summarised in Figure \ref{fig:Training_Diagram}. 

For all models, there was a list of potential hyperparameters that we could have chosen relating to the features of each model architecture. To decide which were best, we trained all possible sets of hyperparameters from the list, recorded the validation loss for the years 2020 and 2022, and chose the hyperparameter set that optimised the validation of the model for the rest of our analysis. Neither of these years is used as test years when evaluating our models. The hyperparameters tested are summarised in Table \ref{tab:Hyperparameters_table}.

\begin{table}[h]
\centering
\resizebox{\textwidth}{!}{%
\begin{tabular}{|>{\centering\arraybackslash}m{0.15\linewidth}|>{\centering\arraybackslash}m{0.22\linewidth}|>{\centering\arraybackslash}m{0.16\linewidth}|>{\centering\arraybackslash}m{0.13\linewidth}|>{\centering\arraybackslash}m{0.15\linewidth}|>{\centering\arraybackslash}m{0.12\linewidth}|}
\hline
\textbf{Model} & \textbf{Hidden Size}  &\textbf{Number of Layers} 
& \textbf{Learning Rate} & \textbf{Bidirectional} & \textbf{Dropout} \\ \hline
\textbf{Single-Catchment} & [16,64,\textbf{128}] &[\textbf{1},2,3]
& [\textbf{1e-3}, 1e-5] & [\textbf{No}, Yes]& [\textbf{0}, 0.1, 0.4] \\ \hline
Multi-Catchment: without River Discharge & [64, \textbf{128}, 256] &[1,\textbf{2},3]
& [\textbf{1e-3}, 1e-5] & [\textbf{No}, Yes] & [0, \textbf{0.2}, 0.4] \\ \hline
\textbf{Multi-Catchment: with River Discharge} & [64, \textbf{128}, 256] &[1,\textbf{2},3]
& [\textbf{1e-3}, 1e-5] & [\textbf{No}, Yes] & [0, \textbf{0.2}, 0.4] \\ \hline
\textbf{Flag LSTM} & [64, \textbf{128}, 256] &[1,\textbf{2},3]
& [\textbf{1e-3}, 1e-5] & [\textbf{No}, Yes] & [0, \textbf{0.2}, 0.4] \\ \hline
\textbf{Hydra-LSTM} & 
    \begin{tabular}[c]{@{}c@{}} \textbf{Body:}  [64, \textbf{128}, 256] \\ \textbf{Head:} [16, \textbf{32}, 64] \end{tabular}  & \begin{tabular}[c]{@{}c@{}} \textbf{Body:}  [1, \textbf{2}, 3] \\ \textbf{Head:} [\textbf{1}, 2] \end{tabular}& 
    [\textbf{1e-3}, 1e-5] & [\textbf{No}, Yes] & [\textbf{0}, 0.2, 0.4] \\ \hline
\end{tabular}
}
\caption{Hyperparameters tested for each model architecture. The hyperparameters that were found to minimize the quantile loss in the validation dataset are in bold. The hidden size relates to the number of parameters in each layer of the LSTM, while the number of layers determines the number of processing steps required to transform the inputs into a satisfactory prediction. The learning rate is a determiner of how much to calibrate the parameters of the model given a particular set of training examples. Bidirectionality determines whether or not the model can use information on what happens in subsequent days to inform its representation of a previous day, and can be useful in some complex problems. Finally, dropout is a process by which parameters in the model are randomly turned off and can often prevent an over-reliance on particular features. Models are color-coded throughout the figures as follows: Single-Catchment LSTMs (light blue), Multi-Catchment LSTM without River Discharge (pink), Multi-Catchment LSTM with River Discharge (purple), Flag LSTM (orange), and Hydra-LSTM (yellow).}
\label{tab:Hyperparameters_table}
\end{table}

\clearpage

\section{Results}

\subsection{Models without River Discharge available as input}
\begin{figure}[h]
    \centering
    \includegraphics[width=0.7\textwidth]{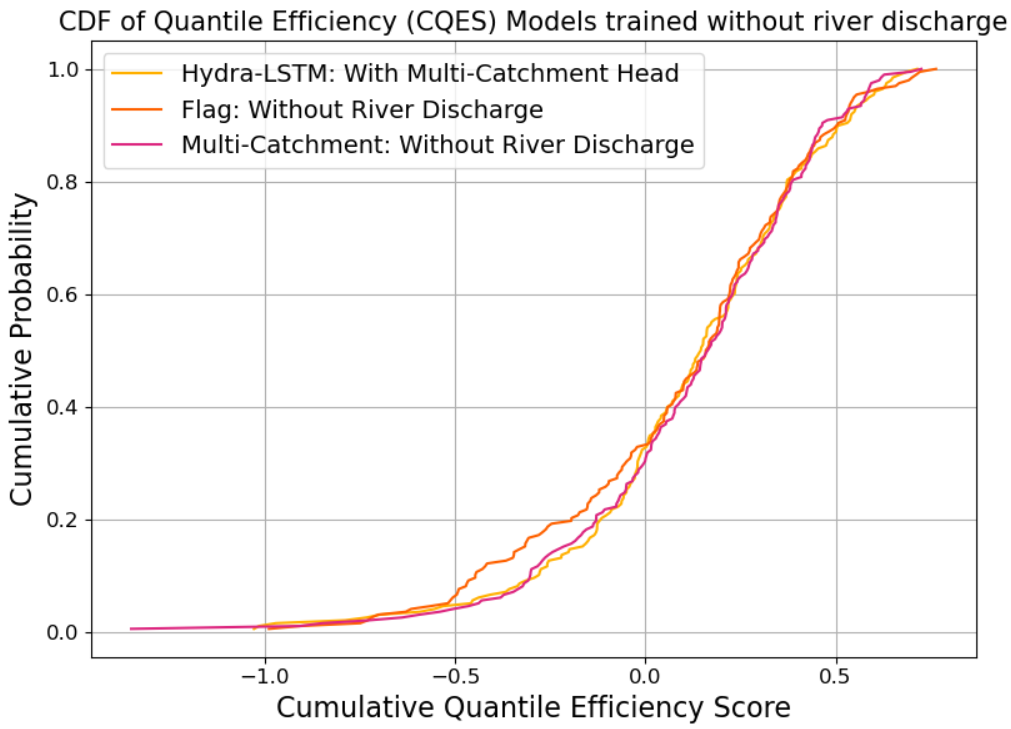}
    \caption{Cumulative Distribution Plot showing the range of Cumulative Quantile Efficiency scores (CQES) \ref{eq: Cumulative Quantile Efficiency Score}, for each model trained without River Discharge as an input. Each individual score is for a single year in a single basin.}
    \label{fig:CDF_Without_Discharge}
\end{figure}

The Multi-Catchment LSTM trained without river discharge as an input performs nearly identically to the Hydra-LSTM with a Multi-Catchment Head, both having a CQES of 0.12 and having empirical 10\% and 90\% quantile thresholds within 1\% of each other. The scores can be found in Table \ref{tab:comparison_without_discharge} and Figure \ref{fig:CDF_Without_Discharge}. This is expected, as the Hydra Body is trained only to minimise the results from the Multi-Catchment Head, and so using the Multi-Catchment Head is akin to using a Multi-Catchment LSTM without river discharge. The Flag LSTM performs slightly worse than the other models in the ungauged setting, with a CQES of 0.09, and its 90\% quantile threshold prediction being exceeded only 85\% of the time compared to 87\% and 88\% of the time for the Multi-Catchment LSTM and Hydra-LSTM. The distribution of CQES scores (Figure \ref{fig:CDF_Without_Discharge}) shows that the distribution of scores for all the models where river discharge is assumed to be unavailable is very similar, with all three curves mostly overlapping. Between a CQES of -0.5 and 0.0 we see the performance of the Flag LSTM being worse than that of the other models, with the curve lying left of the other models.

\begin{table}[h]
    \centering
    \begin{tabular}{|>{\centering\arraybackslash}m{0.2\linewidth}|>{\centering\arraybackslash}m{0.2\linewidth}|>{\centering\arraybackslash}m{0.24\linewidth}|>{\centering\arraybackslash}m{0.24\linewidth}|}
        \hline
        \textbf{Model} &
        \textbf{Cumulative Quantile Efficiency Score} & 
        \textbf{Proportion of Observations exceeding predicted 10\% Quantile Threshold } &  
        \textbf{Proportion of Observations exceeding predicted 90\% Quantile Threshold } \\
        \hline
        \textbf{Multi-Catchment: Without River Discharge} & \textbf{0.12} & 0.13 & 0.87 \\
        \hline
        \textbf{Flag} & 0.09 & 0.13 & 0.85 \\
        \hline
        \textbf{Hydra-LSTM: Multi-Catchment Head} & \textbf{0.12} & \textbf{0.12} & \textbf{0.88} \\
        \hline
    \end{tabular}
    \caption{Comparison of models without river discharge as an available input. Models are color-coded throughout the figures as follows: Single-Catchment LSTMs (light blue), Multi-Catchment LSTM without River Discharge (pink), Flag LSTM (orange), and Hydra-LSTM (yellow). Metrics shown are the Cumulative Quantile Efficiency Score (CQES), the proportion of observations exceeding the models' predicted 10\% quantile threshold, and the proportion of observations exceeding the models' predicted 90\% quantile threshold.}
    \label{tab:comparison_without_discharge}
\end{table}

\subsection{Models with River Discharge available as an input}

\begin{figure}[h]
    \centering
    \includegraphics[width=0.7\textwidth]{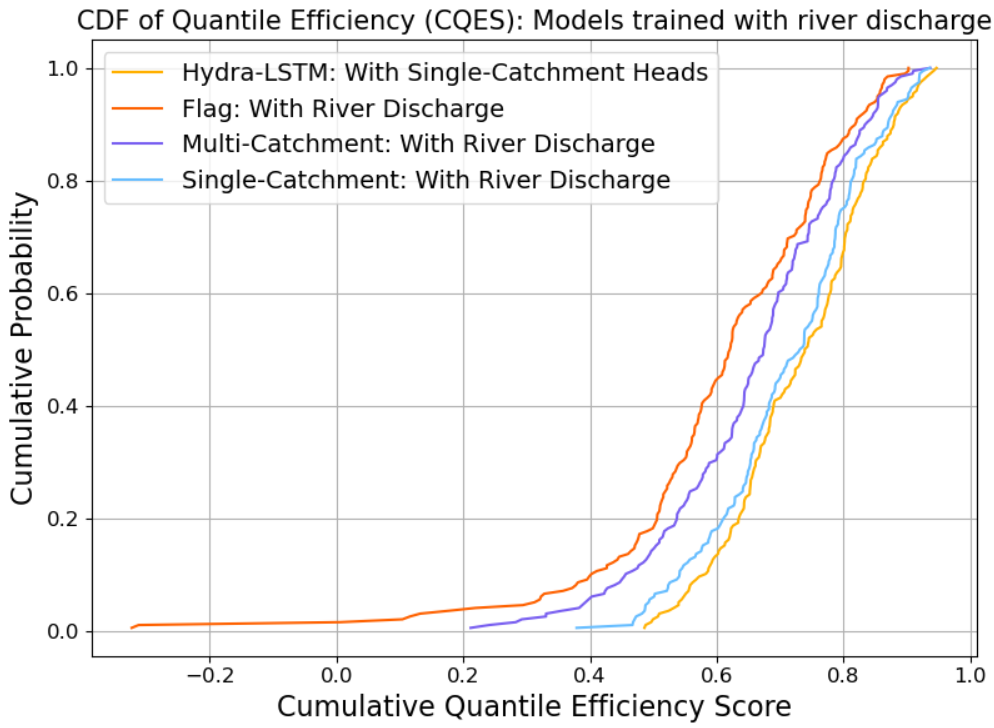}
    \caption{Cumulative Distribution Plot showing the range of Cumulative Quantile Efficiency scores (CQES) \ref{eq: Cumulative Quantile Efficiency Score}, for each model trained with River Discharge as an input. Each individual score is for a single year in a single basin. The CQES scores for the Hydra-LSTM are significantly higher than that of the other models (p = 0.005)}
    \label{fig:CDF_With_Discharge}
\end{figure}

When river discharge is available as an input to each of these models, introduced into the Hydra-LSTM through the use of Single-Catchment Heads, the average skill of each model is drastically increased, with the minimum CQES now being 0.61 compared to a previous maximum of 0.12. The scores in this setting can be found in Table \ref{tab:comparison_with_discharge} and cumulative CQES distribution plot in Figure \ref{fig:CDF_With_Discharge}. The Hydra-LSTM with Single-Catchment Heads has the highest CQES, and the Flag LSTM still has the worst score in the context where river discharge is a usable input. The Single-Catchment LSTMs produce an 80\% confidence interval that is too narrow on average, with its 10\% quantile threshold in actuality being exceeded 15\% of the time and its 90\% quantile threshold only being exceeded 84\% of the time. All other models have empirical quantiles within 3\% of their target. Overall, we find that the Hydra-LSTM with Single-Catchment Heads performs the best of all the models trained with river discharge as an available output. The distribution of average CQES scores for the Hydra-LSTM with Single-Catchment head is the highest compared to all the other models, being the rightmost curve with a lowest CQES score of 0.5 compared to the next best lowest score of below 0.4 for the Single Catchment LSTM. The Hydra-LSTM with Single-Catchment Heads has an average CQES of 0.73 compared to 0.71 for the Single-Catchment LSTM, all other models score lower. The Hydra-LSTM with Single Catchment Heads also had the joint most accurate 10\% quantile threshold estimate, alongside the Flag LSTM, as can be seen in Table \ref{tab:comparison_with_discharge}.

\begin{table}[h]
    \centering
    \begin{tabular}{|>{\centering\arraybackslash}m{0.2\linewidth}|>{\centering\arraybackslash}m{0.2\linewidth}|>{\centering\arraybackslash}m{0.24\linewidth}|>{\centering\arraybackslash}m{0.24\linewidth}|}
        \hline
        \textbf{Model} &
        \textbf{Cumulative Quantile Efficiency Score} & 
        \textbf{Proportion of Observations exceeding predicted 10\% Quantile Threshold } &  
        \textbf{Proportion of Observations exceeding predicted 90\% Quantile Threshold } \\
        \hline
        \textbf{Single-Catchment} & 0.71 & 0.15 & 0.84 \\
        \hline
        \textbf{Multi-Catchment: With River Discharge} & 0.66 & 0.07 & \textbf{0.91} \\
        \hline
        \textbf{Flag} & 0.61 & \textbf{0.09} & 0.92 \\
        \hline
        \textbf{Hydra-LSTM: Single-Catchment Heads} & \textbf{0.73} & \textbf{0.11} & 0.88 \\
        \hline
    \end{tabular}
    \caption{Comparison of models with river discharge as an available input. Models are color-coded throughout the figures as follows: Single-Catchment LSTMs (light blue), Multi-Catchment LSTM with River Discharge (purple), Flag LSTM (orange), and Hydra-LSTM (yellow). Metrics shown are the Cumulative Quantile Efficiency Score (CQES), the proportion of observations exceeding the models' predicted 10\% quantile threshold, and the proportion of observations exceeding the models' predicted 90\% quantile threshold.}
    \label{tab:comparison_with_discharge}
\end{table}

\newpage

\subsection{Case Study: Green River 2001}

\begin{figure}[h!]
    \centering
    \includegraphics[width=0.8\textwidth]{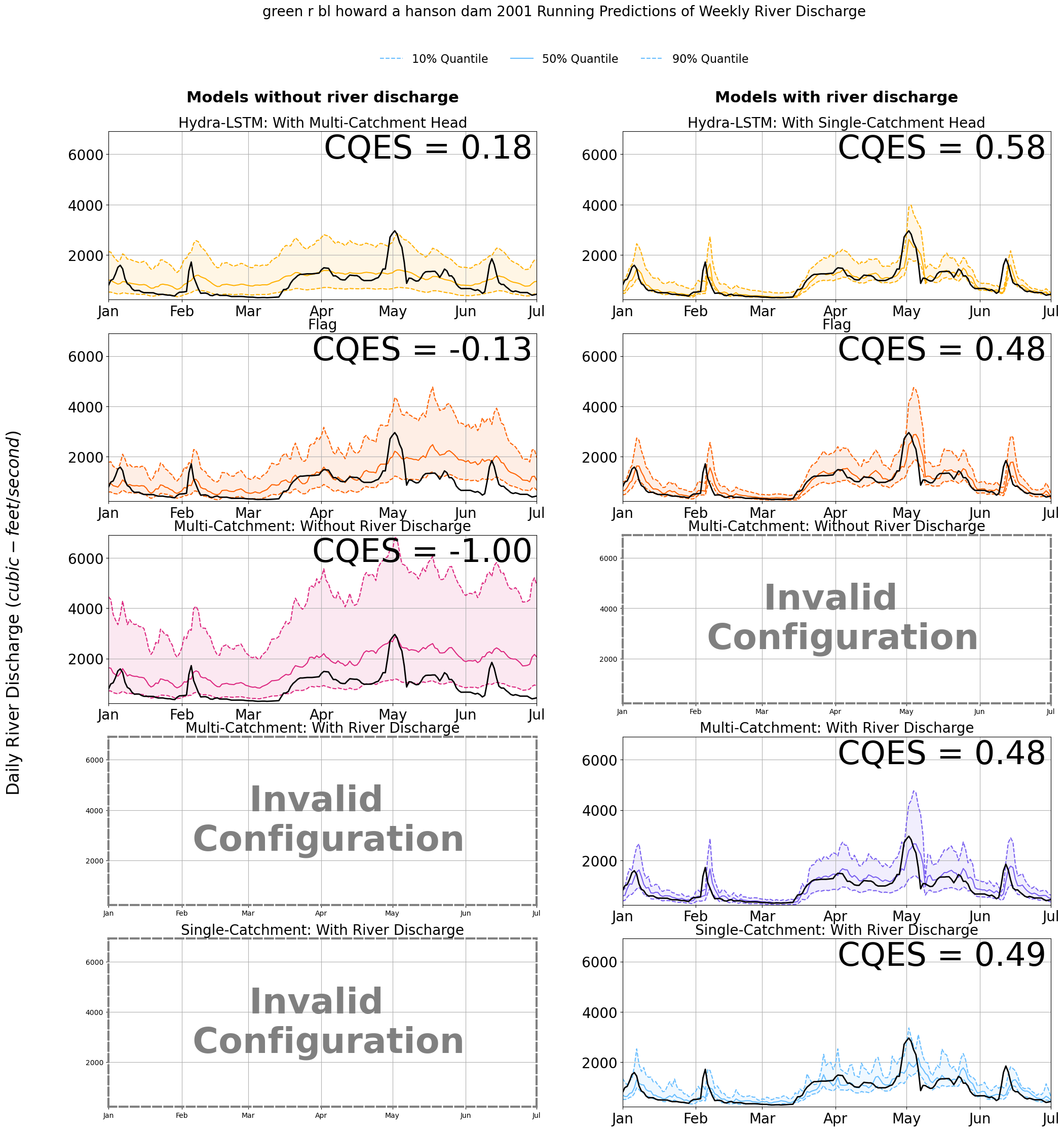}
    \caption{Comparison of the hydrographs of all models for Green River Howard A Hanson Dam, USGS ID 12105900, in 2001. Not all models are usable with all potential data sets, and in this case `Invalid Configuration' is written on the graph. The left column shows how each model performs without river discharge as a potential input, and the right column shows the models with river discharge as a potential input.}  
    \label{fig:Hydrographs}
\end{figure}

In order for hydrologists to be able to reliably use a model, they require assurances that the model will not behave poorly in unseen, or out-of-distribution, conditions. In physical models this can come from much of the physics being prescribed in model development, which is not true in purely statistical or machine learning models. Instead, to test the ability of our models to create skillful predictions in unseen conditions, we closer analyse the January to July river flow in Green River,  USGS ID 12105900, for 2001 (Figure \ref{fig:Hydrographs}). Green River saw particularly low flow in 2001,  with a peak flow roughly three times less than the average flow in other years. The Hydra-LSTM was best able to capture the potential for high flow being much lower than in other years, with the predicted 90\% quantile thresholds being much lower than in other years. In comparison, the other models still had relatively high 90\% quantile thresholds. The Multi-Catchment LSTM trained without river discharge performed the worst at this catchment, with its 90\% quantile threshold remaining much closer to the usual flow seen in that catchment in other years, of approximately 6000 cubic feet per second as opposed to the 2000 feet per second seen in that year. 

The Hydra-LSTM with Single-Catchment Heads also had the best CQES score in the gauged setting for this scenario, as seen in the right column of Figure \ref{fig:Hydrographs}. Again, it seems that the Flag LSTM and Multi-Catchment LSTM overestimated the 10\% quantile threshold, and this is confirmed when looking at the proportion of 10\% quantile thresholds that were exceeded in Table \ref{tab:comparison_with_discharge}. The Single-Catchment LSTM, on the other hand, seemed to have an overly constrained 10\% quantile threshold prediction, being exceeded 15\% of the time. The corresponding hydrograph shows a high degree of stochasticity in its daily predictions, significantly less smooth than the true observations.

\newpage

\section{Discussion}

\subsection{Hydra-LSTM Multi-Catchment Head at least as skillful as other state-of-the-art machine learning models}
This paper develops a probabilistic model to match existing state-of-the-art architectures while allowing individual forecasters to introduce new inputs after the initial development of the model. We did this by developing a model with three key components: a Hydra Body to transform data available across all catchments into something more directly useful for river discharge prediction, a Multi-Catchment Hydra Head to take the transformations from the Hydra Body and make quantile predictions for river discharge when additional data is not available; and a suite of Single-Catchment Hydra Heads to take in additional data available only at select catchments, and process these alongside the transforms produced by the Hydra Body to make more informed predictions when possible. 

In the first setting, when no additional data is available, and only the Hydra Body and Multi-Catchment Hydra head are used, we have shown that the performance of the Hydra-LSTM is on par with other state-of-the-art machine learning models used in river discharge prediction, namely the Multi-Catchment LSTM (Kratzert et al. 2019) and the Flag LSTM (Nearing et al. 2023). Not only does it have the best cumulative score, as defined in equation \ref{eq: Cumulative Quantile Efficiency Score} and shown in Table \ref{tab:comparison_without_discharge} and Figure \ref{fig:CDF_Without_Discharge}, but it also has the most accurate quantile thresholds. The Hydra-LSTM in this setting is expected to have results nearly identical to the Multi-Catchment LSTM, as the Hydra Body is trained only to minimise the loss from the Multi-Catchment Hydra Head. This means that the combination is the same as a two layer LSTM with a different hidden size for each layer. It may also be expected that the Flag LSTM would perform slightly worse, as it attempts to create predictions both when historic river discharge is available as an input and when it is not, without having an auxiliary head as the Hydra-LSTM can have. 

\subsection{Hydra-LSTM Single-Catchment Heads  at least as skillful as effective at using additional variables as if it were introduced in the initial set}
The training of the Hydra-LSTM Single-Catchment Heads is very efficient, as it needs only to be a single layer LSTM, using the transformations learned by the Hydra Body as the main set of inputs. Using parallel processes on an A100 GPU, we could train 11 different folds of each of the models in under 2 hours and train 11 different single-catchment heads in under an hour. Because of its efficiency, we do not see the required resources to train an additional Single-Catchment Hydra Head as a significant computational burden for most users. The efficacy of the Single-Catchment Hydra Head is also evident from our results, with the distribution of scores lying strongly to the right, meaning higher scores than that of the other models, Figure \ref{fig:CDF_With_Discharge}. It had a significantly higher CQES score of 0.73 than the Flag LSTM, which had the lowest score of 0.61 of all the models, and its quantile thresholds were within exceeded 2\% of their intended values as seen in Table \ref{tab:comparison_with_discharge}. We believe that the significant improvements compared to the Flag LSTM, which is the only other model that can be used with variable data available, shows that it may be a better alternative in many cases, for example, when the additional data available is only available for a single catchment. This is because the Flag LSTM and the Multi-Catchment LSTM with river discharge were trained with river discharge across all catchments, and so were trained on more data than might be available for another variable we might wish to introduce, such as radar measurements were taken only at a single catchment or upstream information for a particular catchment. It may be that changing the proportion of training devoted to a variable being available as an input and not available as an input might have improved the Flag LSTM, however this is an additional complexity decreasing the usability of the Flag LSTM. Training a Flag LSTM or Multi-Catchment would also require the forecaster to have access to and train the model on all the relevant data for all the catchments the model was trained on, whereas to train an additional Single-Catchment Head in the Hydra-LSTM the forecaster would only need to train on the data for their own catchments, as the Hydra-Body does not need to be retrained on each catchment.

\subsection{Hydra-LSTM Single-Catchment Heads offer better performance than training a new Single-Catchment LSTM}
Not only does the Hydra-LSTM outperform the other models that are trained across multiple catchments, but we have also shown that it, on average, outperforms the use of different LSTMs at each catchment. Table \ref{tab:comparison_with_discharge} shows that the Single-Catchment LSTM, while having the second highest average CQES, tends to underestimate the uncertainty in its forecasts, with an average interval between its 10\% and 90\% quantile thresholds of 69\%, compared to one of 77\% for the Hydra-LSTM. The Hydrographs in Figure \ref{fig:Hydrographs} suggest that Single-Catchment LSTMs may over-rely on persistence. Its predictions for the 10\% quantile threshold also seemed to be particularly erratic, having a high daily variability that was much more than seen in other models or in the actual flow in  the catchment. that is more than is seen in this catchment. This may suggest an inability to fully appreciate some of the underlying hydrological behaviour, having been trained on the data on only one catchment as opposed to being driven by a model trained over many catchments, as the Hydra Body is. 

Overall, this suggests that the Hydra-LSTM is just as skilful a method as the other models when there is no additional data available that a local forecaster would wish to add as a predictor for their catchment, and that if desired, a forecaster can introduce additional data in a Single-Catchment Head in a way that is just as effective as retraining a Multi-Catchment LSTM.

\subsection{Aspects for Future Work}

Our results have shown the ability of the Hydra-LSTM to introduce catchment-specific data, and it would be valuable for future work to assess how much historical data is needed for the Single-Catchment Heads to learn from this additional data. The Hydra Body may allow for additional skill to be gained from additional data that has too short a historical record to be used to train a Single-Catchment LSTM. 

Further work may wish to introduce forecast data as a potential input. In this paper, we did not include forecast data as inputs to our model; however, for longer lead times, it will be useful to introduce the predictions from atmospheric forecast models such as the ECMWF IFS (Persson and
Grazzini 2007).  This addition can be introduced into the Hydra-LSTM architecture, for example, by using the Hindcast-Forecast LSTM developed by \cite{nearing2023ai} as the model blocks. 

It is also possible to use flag indicators in the Hydra-LSTM if there are variables that might be best introduced in the Hydra Body with a flag indicator instead of as a separate head. Future research may wish to explore how these two means of introducing data optionally can be used together and when it is worth rebuilding the Hydra Body instead of introducing additional data in the Single-Catchment Head. Flags have the unique advantage of allowing us to use them still when data is only intermittently available. In contrast, the Single-Catchment Heads have the advantage of being able to introduce new variables without having to retrain the entire model. Testing the feasibility of adding flag time series into the Hydra Model would allow us to take full advantage of different data types at each catchment.

Another strength of the Hydra-LSTM that future work should explore is that it can have multi-catchment Heads that are useful for a subset of catchments that the multi-catchment head is useful for, but more than for a single catchment. In our experiments, we trained a different Single-Catchment Head to introduce gauge river discharge as an input for each catchment, and this was necessary in order to test the ability for variables to be incorporated even when we only had the data for one catchment to train on. It would also have been possible, however, to train one other Multi-Catchment Head that combined the outputs from the Hydra Body with historic river discharge and was trained over all catchments that had river discharge. This may be worth doing when not all the catchments we want our models to be used for have a variable available, such as historic river discharge, but enough do that it is worth creating an alternate head and training it on all relevant data across catchments.

Finally, the benefits seen in the Hydra-LSTM are likely to evolve as the number of catchments the Hydra Body can be trained on increases. \cite{kratzert2023never} have shown the performance gains from training along multiple catchments to increase even above hundreds of catchments, and this bonus is likely to be seen in the Hydra Body as well. 

\section{Conclusion}

Before this paper, no method existed that allowed for the easy introduction of new variables outside of an initial set. In this paper, we have developed the Hydra-LSTM, a new machine-learning architecture for predicting river discharge in a multi-catchment setting. This method allows organisations to create large-scale models to make predictions across multiple catchments without restricting them to any particular subset of data to be used as predictors. 
The Hydra-LSTM is equivalent in skill to other state-of-the-art models when a catchment manager does not need to introduce their own variables, and the Single-Catchment Heads of the Hydra-LSTM can introduce new variables into the predictor set in a way that is more effective than introducing in the base set of predictands in other models.  The Single-Catchment Heads being a more effective means of introducing additional variables is a bonus; it is the only realistic option for a local forecaster to add their own variables into a model. We encourage future work to test the Hydra-LSTM to introduce new variables that are specific to individual regions, to increase the number of catchments the Hydra-LSTM is trained on to assess its scaling properties, and to test in cases for which available data is limited.

%%%
% \section{First primary heading}

% \subsection{First secondary heading}

% \subsubsection{First tertiary heading}

% \paragraph{First quaternary heading}

%%%%%%%%%%%%%%%%%%%%%%%%%%%%%%%%%%%%%%%%%%%%%%%%%%%%%%%%%%%%%%%%%%%%%
% TABLES---INSERT NEAR IN-TEXT DISCUSSION
%%%%%%%%%%%%%%%%%%%%%%%%%%%%%%%%%%%%%%%%%%%%%%%%%%%%%%%%%%%%%%%%%%%%%
%%  Enter tables near where they are discussed within the document. 
%%  Please place tables before/after paragraphs, not within a paragraph.
%%
%
%\begin{table}[t]
%\caption{This is a sample table caption and table layout.  Enter as many tables as
%  necessary at the end of your manuscript. Table from Lorenz (1963).}\label{t1}
%\begin{center}
%\begin{tabular}{ccccrrcrc}
%\hline\hline
%$N$ & $X$ & $Y$ & $Z$\\
%\hline
% 0000 & 0000 & 0010 & 0000 \\
% 0005 & 0004 & 0012 & 0000 \\
% 0010 & 0009 & 0020 & 0000 \\
% 0015 & 0016 & 0036 & 0002 \\
% 0020 & 0030 & 0066 & 0007 \\
% 0025 & 0054 & 0115 & 0024 \\
%\hline
%\end{tabular}
%\end{center}
%\end{table}

%%%%%%%%%%%%%%%%%%%%%%%%%%%%%%%%%%%%%%%%%%%%%%%%%%%%%%%%%%%%%%%%%%%%%
% FIGURES---INSERT NEAR IN-TEXT DISCUSSION
%%%%%%%%%%%%%%%%%%%%%%%%%%%%%%%%%%%%%%%%%%%%%%%%%%%%%%%%%%%%%%%%%%%%%
%%  Enter figures near where they are discussed within the document.
%%  Please place figures before/after paragraphs, not within a paragraph.
% %
%
%\begin{figure}[t]
%  \noindent\includegraphics[width=19pc,angle=0]{figure01.pdf}\\
%  \caption{Enter the caption for your figure here.  Repeat as
%  necessary for each of your figures. Figure from \protect\cite{Knutti2008}.}\label{f1}
%\end{figure}

\clearpage
%%%%%%%%%%%%%%%%%%%%%%%%%%%%%%%%%%%%%%%%%%%%%%%%%%%%%%%%%%%%%%%%%%%%%
% ACKNOWLEDGMENTS
%%%%%%%%%%%%%%%%%%%%%%%%%%%%%%%%%%%%%%%%%%%%%%%%%%%%%%%%%%%%%%%%%%%%%
\acknowledgments

This work was supported by the Advanced Frontiers for Earth System Prediction Doctoral Training Programme, funded by the University of Reading.

%  Keep acknowledgments (note correct spelling: no ``e'' between the ``g'' and
% ``m'') as brief as possible. In general, acknowledge only direct help in
%  writing or research. Financial support (e.g., grant numbers) for the work done, 
%  for an author, or for the laboratory where the work was performed must be 
%  acknowledged here rather than as footnotes to the title or to an author's name.
%  Contribution numbers (if the work has been published by the author's institution 
%  or organization) should be placed in the acknowledgments rather than as 
%  footnotes to the title or to an author's name.

%%%%%%%%%%%%%%%%%%%%%%%%%%%%%%%%%%%%%%%%%%%%%%%%%%%%%%%%%%%%%%%%%%%%%
% DATA AVAILABILITY STATEMENT
%%%%%%%%%%%%%%%%%%%%%%%%%%%%%%%%%%%%%%%%%%%%%%%%%%%%%%%%%%%%%%%%%%%%%
% 
%
\datastatement

%  The data availability statement is where authors should describe how the data underlying 
%  the findings within the article can be accessed and reused. Authors should attempt to 
%  provide unrestricted access to all data and materials underlying reported findings. 
%  If data access is restricted, authors must mention this in the statement. See
%  {http://www.ametsoc.org/PubsDataPolicy} for more info.

%%%%%%%%%%%%%%%%%%%%%%%%%%%%%%%%%%%%%%%%%%%%%%%%%%%%%%%%%%%%%%%%%%%%%
% APPENDIXES
%%%%%%%%%%%%%%%%%%%%%%%%%%%%%%%%%%%%%%%%%%%%%%%%%%%%%%%%%%%%%%%%%%%%%
%
%% If only one appendix, use

%\appendix

%% If more than one appendix, use \appendix[<letter>], e.g.,

%\appendix[A] 

%% Appendix title is necessary! For appendix title:

%\appendixtitle{Title of Appendix}

%%% Appendix section numbering (note, skip \section and begin with \subsection)
%
% \subsection{First primary heading}

% \subsubsection{First secondary heading}

% \paragraph{First tertiary heading}

%%%%%%%%%%%%%%%%%%%%%%%%%%%%%%%%%%%%%%%%%%%%%%%%%%%%%%%%%%%%%%%%%%%%%
% REFERENCES
%%%%%%%%%%%%%%%%%%%%%%%%%%%%%%%%%%%%%%%%%%%%%%%%%%%%%%%%%%%%%%%%%%%%%
% Make your BibTeX bibliography by using these commands:
\bibliographystyle{ametsocV6}
\bibliography{references}

\end{document}